\documentclass[10pt,twocolumn,letterpaper]{article}

\usepackage{iccv}
\usepackage{times}
\usepackage{epsfig}
\usepackage{graphicx}
\usepackage{amsmath}
\usepackage{amssymb}

\usepackage[pagebackref=true,breaklinks=true,letterpaper=true,colorlinks,bookmarks=false]{hyperref}

\iccvfinalcopy 

\def\iccvPaperID{667} 

\ificcvfinal\fi

\usepackage{paralist}
\usepackage{bm}
\usepackage[american]{babel}

\begin{document}

\title{VQS: Linking Segmentations to Questions and Answers for \\ Supervised Attention in VQA and Question-Focused Semantic Segmentation}

\author{Chuang Gan$^{1}$  \quad  Yandong Li$^{2}$ \quad Haoxiang Li$^{3}$ \quad  Chen Sun$^{4}$ \quad Boqing Gong$^{2}$ \\
$^1$IIIS, Tsinghua University, China \quad
$^2$CRCV, University of Central Florida, USA\\
$^3$Adobe Research, USA \quad
$^4$Google Research, USA \\
 }

\maketitle

\begin{abstract}
   Rich and dense human labeled datasets are among the main enabling factors for the recent advance on vision-language understanding. Many seemingly distant annotations (e.g., semantic segmentation and visual question answering (VQA)) are inherently connected in that they reveal different levels and perspectives of human understandings about the same visual scenes --- and even the same set of images (e.g., of COCO). The popularity of COCO correlates those annotations and tasks. Explicitly linking them up may significantly benefit both individual tasks and the unified vision and language modeling.

    We present the preliminary work of linking the instance segmentations provided by COCO to the questions and answers (QAs) in the VQA dataset, and name the collected links \emph{visual questions and segmentation answers (VQS)}. They transfer human supervision between the previously separate tasks, offer more effective leverage to existing problems, and also open the door for new research problems and models. We study two applications of the VQS data in this paper: supervised attention for VQA and a novel question-focused semantic segmentation task. For the former, we obtain state-of-the-art results on the VQA real multiple-choice task by simply augmenting the multilayer perceptrons with some attention features that are learned using the segmentation-QA links as explicit supervision. To put the latter in perspective, we study two plausible methods and compare them to an oracle method assuming that the instance segmentations are given at the test stage. 

\vspace{-8mm}
\end{abstract}
\let\thefootnote\relax\footnote{Code and data: {https://github.com/Cold-Winter/vqs}.}





\section{Introduction}
\label{sec:intro}

Connecting visual understanding with natural language has received extensive attentions in recent years. We have witnessed the resurgence of image captioning~\cite{show_and_tell,mao2014deep,karpathy2015deep,LRCN,sun2015concept, Mideye,attend,gan2017semantic,pan2016jointly,gan2017stylenet} which is often addressed by jointly modeling visual and textual content with deep neural networks. However, image captions tend to be diverse and subjective --- it is hard to evaluate the quality of captions generated by different algorithms~\cite{elliott2014comparing,cider,spice}, and tend to miss subtle details --- in training, the models may be led to capturing the scene-level gist rather than fine-grained entities. In light of the premises and demerits of image captioning, visual question answering (VQA)~\cite{VQA,visual7w,COCOQA,vqa_baidu} and visual grounding~\cite{Flickr30k_entities,segmentation_gronding,Grounding,comprehension,NLP_object_retrieval,wang2016learning,yu2016modeling} are proposed, in parallel, to accommodate automatic evaluation and multiple levels of focus on the visual entities (e.g., scene, object, activity, attribute, context, relationships, etc.).

\begin{figure}[t]
   \centering
   \includegraphics[width = 1\linewidth]{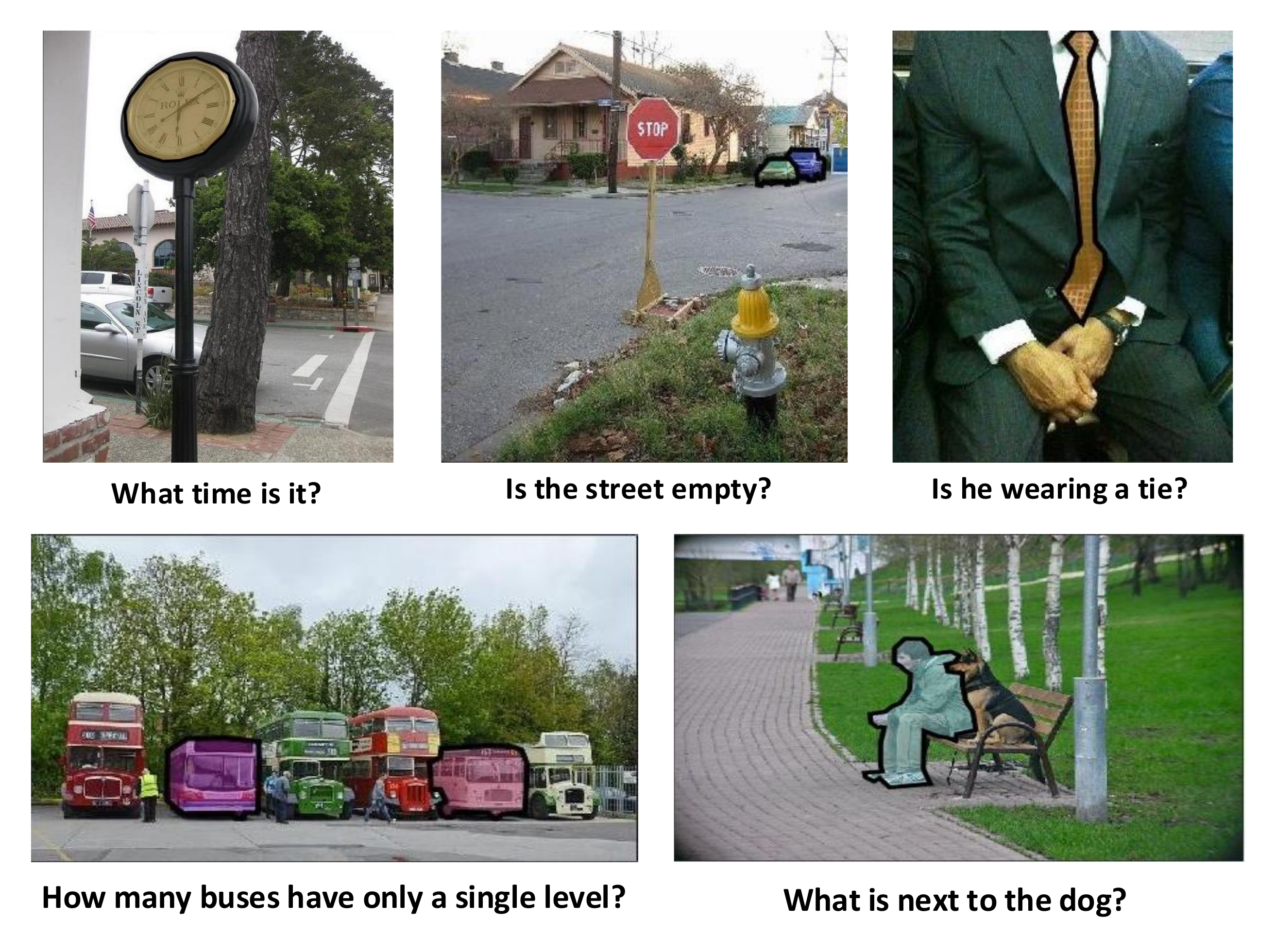}

   \caption{Taking as input an image and a question about the image, an algorithm for the question-focused semantic segmentation is desired to generate some segmentation mask(s) over the entities in the image that can visually answer the question.}
   \label{fig:1}
   \vspace{-15pt}
\end{figure}

Rich and dense human annotated datasets are arguably the main ``enabler'', among others, for this line of exciting works on vision-language understanding. COCO~\cite{COCO} is especially noticeable among them. It contains mainly classical labels (e.g., segmentations, object categories and instances, key points, etc.) and image captions. Many research groups have then collected additional labels of the COCO images for a variety of tasks. Agrawal et al.\ crowdsourced questions and answers (QAs) about a subset of the  COCO images and abstract scenes~\cite{VQA}. Zhu et al.\ collected seven types of QAs in which the object mentions are associated with bounding boxes in the images~\cite{visual7w}. Mao et al.~\cite{comprehension} and Yu et al.~\cite{yu2016modeling} have users to give referring expressions that each pinpoints a unique object in an image. The Visual Genome dataset~\cite{krishna2016visual} also intersects with  COCO in terms of images and provides dense human annotations, especially scene graphs.

These seemingly distant annotations are inherently connected in the sense that they reveal different perspectives of human understandings about the same  COCO images. The popularity of  COCO could strongly correlate those annotations --- and even tasks. Explicitly  linking them up, as we envision, can significantly benefit both individual tasks and unified vision-language understanding, as well as the corresponding approaches and models. One of our contributions in this paper is to initiate the preliminary work on this. In particular, we focus on linking the segmentations provided by COCO~\cite{COCO} to the QAs in the VQA dataset~\cite{VQA}. Displaying an image and a QA pair about the image, we ask the participant to choose the segmentation(s) of the image in order to visually answer the question.

Figure~\ref{fig:1} illustrates some of the collected ``visual answers''. For the question ``What is next to the dog?'', the output is supposed to be the segmentation mask over the man. For the question ``What time is it?'', the clock should be segmented out. Another intriguing example is that the cars are the desired segmentations to answer ``Is this street empty?'', providing essential visual evidence for the simple text answer ``no''. Note that while many visual entities could be mentioned in a question, we only ask the participants to choose the target segmentation(s) that visually answer the question. This simplifies the annotation task and results in higher agreement between participants. Section~\ref{dataset} details the annotation collection process and statistics.

\vspace{-4mm}
\paragraph{Two related datasets.}
Das et al.\ have collected some human attention maps for the VQA task~\cite{humman_attention}. They blur the images and then ask users to scratch them to seek visual cues that help answer the questions. The obtained attention maps are often small, revealing meaningful parts rather than complete objects. The object parts are also mixed with background areas and with each other.  As a result, the human attention maps are likely less accurate supervision for the attention based approaches to VQA than the links we built between segmentations and QAs. Our experiments verify this hypothesis (cf.\ Section~\ref{sVQA}). While bounding boxes are provided in Visual7W~\cite{visual7w} for object mentions in QAs, they do not serve for the purpose of directly answering the questions except for the ``pointing'' type of questions. In contrast, we provide direct visual answers in the form of segmentations to more question types.

\vspace{-1mm}
\subsection{Applications of the segmentation-QA links}
We call the collected links between the COCO segmentations~\cite{COCO} and QA pairs in the VQA dataset~\cite{VQA} \emph{visual questions and segmentation answers (\textbf{VQS})}. {Such links transfer human supervision between the previously separate  tasks, i.e., semantic segmentation and VQA. They enable us to tackle existing problems with more effective leverage than before and also open the door for new research problems and models for the vision-language understanding.} We study two applications of our VQS dataset in this paper:  supervised attention for VQA and a novel question-focused semantic segmentation (QFSS) task. For the former, we obtain state-of-the-art results on the VQA real multiple-choice task by simply augmenting the multilayer perceptrons (MLP) of~\cite{Revisiting_vqa} with attention features.

\vspace{-3mm}
\subsubsection{Supervised attention for VQA}
VQA is designed to answer natural language questions about images in the form of short texts. The attention scheme is often found useful for VQA, by either attending particular image regions~\cite{stacked,xu2015ask,xiong2016dynamic,coattention,li2016visual} or modeling object relationships~\cite{Andreas2016Neural,Multi_World}. However, lacking explicit attention annotations, the existing methods opt for latent variables and use indirect cues (e.g., text answers) for inference. As a result, the machine-generated attention maps are poorly correlated with human attention maps~\cite{humman_attention}. This is not surprising since latent variables hardly match semantic interpretations due to the lack of explicit training signals; similar observations exist in other studies, e.g., object detection~\cite{felzenszwalb2010object}, video recognition~\cite{gan2015devnet} and text processing~\cite{yu2009learning}.

These phenomena highlight the need for explicit links between the visual and text answers, realized in this work as VQS. We show that, by supervised learning to attend different image regions using the collected segmentation-QA links, we can boost the simple MLP model~\cite{Revisiting_vqa} to very compelling performance on the VQA real multi-choice task.

\begin{figure*}[t]

   \centering
   \includegraphics[width = 1\linewidth]{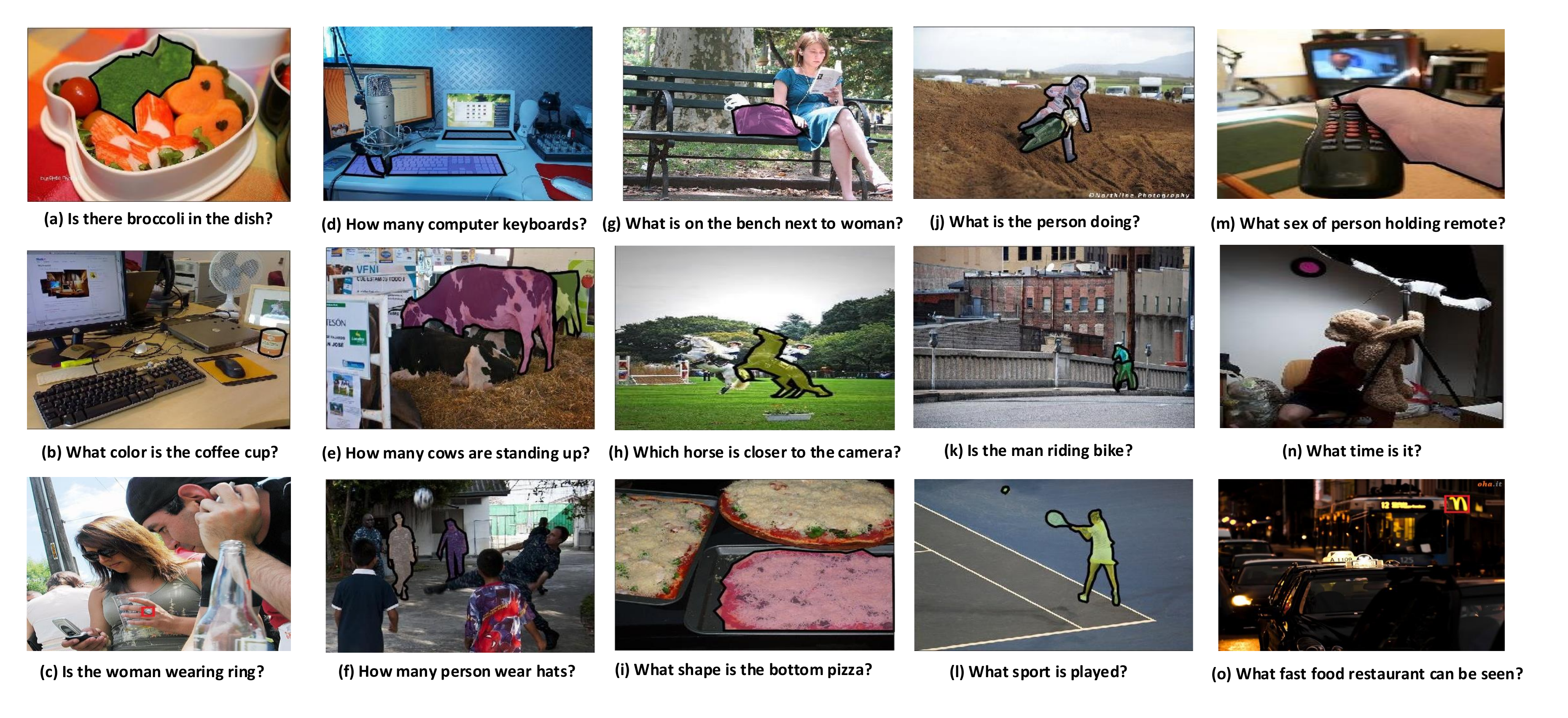}
   \vspace{-6mm}
   \caption{Some typical examples in our VQS dataset. From the left to right, the underlying tasks are respectively about object localization, semantic segmentation, understanding object relationships,  fine-grained activity localization, and commonsense reasoning.}

   \label{fig:example}
   \vspace{-6pt}
\end{figure*}

\vspace{-3mm}
\subsubsection{Question-focused semantic segmentation (QFSS)}
In addition to the supervised attention for better tackling VQA, VQS also enables us to explore a novel question-focused semantic segmentation (QFSS) task.



Since VQA desires only text answers, there exist potential shortcuts for the learning agent, e.g., to generate correct answers without accurately reasoning the locations and relations of different visual entities. While visual grounding (VG) avoids the caveat by placing bounding boxes~\cite{Flickr30k_entities,Grounding,comprehension,NLP_object_retrieval} or segmentations~\cite{segmentation_gronding} over the target visual entities, the scope of the text expressions in existing VG works is often limited to the visual entities present in the image. In order to bring together the best of VQA and VG, we propose the QFSS task, whose objective is to produce pixel-wise segmentations in order to visually answer the questions about images. It effectively borrows the versatile questions from VQA and meanwhile resembles the design of VG in terms of the pixel-wise segmentations as the desired output.

Given an image and a question about the image, we propose a mask aggregation approach to generating a segmentation mask as the visual answer. Since QFSS is a new task, to put it in perspective, we not only compare the proposed approach to competing baselines but also study an upper-bound method by assuming all instance segmentations are given as oracles at the test stage.

Hu et al.'s work~\cite{segmentation_gronding} is the most related to QFSS. They learn to ground text expressions in the form of image segmentations. Unlike the questions used in this work that are flexible to incorporate commonsense and knowledge bases, the expressive scope of the text phrases in~\cite{segmentation_gronding} is often limited to the visual entities in the associated images.

The rest of this paper is organized as follows. Section~\ref{dataset} details the collection process and analyses of our VQS data. In section~\ref{sVQA}, we show how to use the collected segmentation-QA links to learn supervised attention features and to augement the existing VQA methods. In section~\ref{sQFSS}, we study a few potential frameworks to address the new question-focused semantic segmentation task. Section~\ref{conclusion} concludes the paper.

\section{Linking image segmentations to text QAs}
\label{dataset}
In this section, we describe in detail how we collect the links between the semantic image segmentations and text questions and answers (QAs). We build our work upon the images and instance segmentation masks in  COCO~\cite{COCO} and the QAs in the VQA dataset~\cite{VQA}. The  COCO images are mainly about everyday scenes that contain common objects in their natural contexts, accommodating complex interactions and relationships between different visual entities. To avoid trivial links between the segmentations and QA pairs, we only keep the images that contain at least three instance segmentations in this work. The questions in VQA~\cite{VQA} are diverse and comprehensively cover various parts of an image, different levels of semantic interpretations, as well as commonsense and knowledge bases.

Next, we elaborate the annotation instructions and provide some analyses about the collected dataset.

\subsection{Annotation instructions}

We display to the annotators an image, its instance  segmentations from the COCO dataset, and a QA pair about the image from the VQA dataset. The textual answer is given in addition to the question, to facilitate the participants to choose the right segmentations as the visual answer. Here are the instructions we give to the annotators (cf.\ the supplementary materials for the GUI):

\begin{compactitem}
    \item Please choose the right segmentation(s) in the image to answer the question. Note that the text answer is shown after the question.

    \item A question about the target entities may use other entities to help refer to the target. Choose the target entities only and nothing else (e.g., the purse for ``What is on the bench next to woman?'' in Figure~\ref{fig:example}(g)).

    \item A question may be about an activity. Choose all visual entities involved in the activity. Taking Figure~\ref{fig:example}(j) for instance, choose both the person and motorcycle for the question ``what is the person doing?''.

        \item Sometimes, in addition to the image regions covered by the segmentation masks, you may need other regions to answer the question. To include them, please draw tight bounding box(es) over the region(s).

    \item For the ``How many'' type of questions, the number of selected segments (plus bounding boxes) must match the answer. If the answer is greater than three, it is fine to put one bounding box around the entities being asked in the question.

    \item Please tick the black button under the question, if you think the question has to be answered by the full image.

    \item Please tick the gray button under the question, if you feel the question is ambiguous, or if you are not sure which segment/region to select to answer the question.

\end{compactitem}

\begin{figure}[t]
   \centering
   \includegraphics[width = 0.9\linewidth]{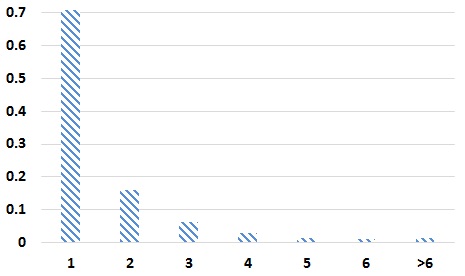}
   \caption{Distribution of the  number of segmentations per question-image pair.}
   \vspace{-10pt}
      \label{fig:segment}
\end{figure}

Occasionally, the visual answer is supposed to be only part of an instance segment given by  COCO. For instance, the McDonald logo answers ``What fast food restaurant can be seen?'' in Figure~\ref{fig:example}(o) but there is no corresponding segmentation for the logo in COCO. Another example is the region of the ring that answers ``Is the woman wearing ring?'' (cf.\ Figure~\ref{fig:example}(c)). For these cases, we ask the participants to draw tight bounding boxes around them. If we segment them out instead, a learning agent for QFSS may never be able to produce the right segmentation for them unless we include more training images in the future, since these regions (e.g., McDonald logo, ring) are very fine-grained visual entities and show up only a few times in our data collection process.

\vspace{-4.5mm}
\paragraph{Quality control.}
We tried AMTurk to collect the annotations at the beginning. While the inter-annotator agreement is high on the questions about objects and people, there are  many inconsistent annotations for the questions referring to activities (e.g., ``What sport is played?''). Besides, the AMTurk workers tend to frequently tick the black button, which says the full image is the visual answer, and the gray button, which tells the question is ambiguous. To obtain higher-quality annotations, we instead invited 10 undergraduate and graduate volunteers and trained them in person (we include some slides used for the training in the supplementary materials). To further control the annotation quality, each annotator was asked to finish an assignment of 100 images (around 300 question-answer pairs) before we met with them again to look over their annotations together --- all the volunteers were asked to participate the discussion and jointly decide the expected annotations for every question.
We also gradually increased the hourly payment rate from \$12/hr to \$14/hr as incentives for high-quality work.
\begin{figure}[t]
   \centering
   \includegraphics[width = 0.85\linewidth]{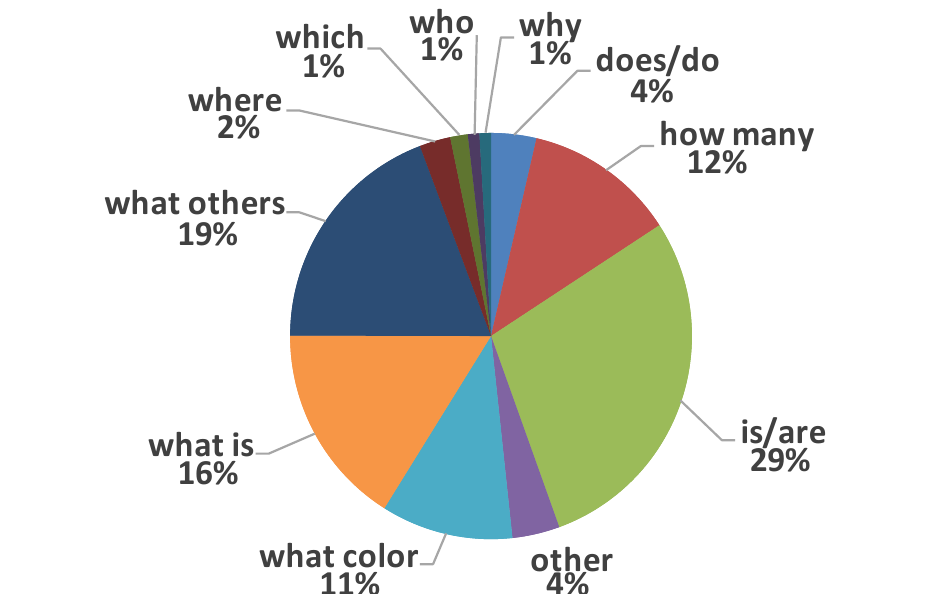}
   \caption{The distribution of question types in the VQS dataset.}
   \vspace{-10pt}
      \label{fig:distrbution}
\end{figure}

\subsection{Tasks addressed by the participants}
Thanks to the rich set of questions collected by Agrawal \textit{et al.}~\cite{VQA} and the complex visual scenes in COCO~\cite{COCO}, the participants have to parse the question, understand the visual scene and context, infer the interactions between visual entities, and then pick up the segmentations that answer the questions. We find that many vision tasks may play roles in this process. Figure~\ref{fig:example} shows some typical examples to facilitate the following discussion.

\textbf{Object detection.} Many questions directly ask about the properties of some objects in the images. In Figure~\ref{fig:example}(b), the participants are supposed to identify the \textit{cup} in the cluttered scene for the question ``What color is the coffee cup?''. 

\textbf{Semantic segmentation.} For some questions, the visual evidence to answers is best represented by semantic segmentations. Take Figures~\ref{fig:example}(j) and (k) for instance. Simply detecting the rider and/or the bike would be inadequate in expressing their spatial interactions.

\textbf{Spatial relationship reasoning.} A question like ``What is on the bench next to the woman?'' (Figure~\ref{fig:example}(g)) poses a challenge to the participants through the spatial relationship between objects including bench, woman, and the answer purse. Figure~\ref{fig:example}(i) is another example in this realm.

\textbf{Fine-grained activity recognition.} When the question is about an activity (\textit{e.g.}, ``What sport is being played?'' in Figure~\ref{fig:example}(l)), we ask the participants to label all the visual entities (\textit{e.g.}, person, tennis racket , and tennis ball) involved. In other words, they are expected to spot the fine-grained details of the activity.


\textbf{Commonsense reasoning.} Commonsense knowledge can help the  participants   significantly reduce the search space for the visual answers, e.g.,  the clock to answer ``What time is it?'' in Figure~\ref{fig:1}, and the McDonald logo to answer ``What fast food restaurant can be seen?'' in Figure~\ref{fig:example}(o).

\begin{figure*}[t]
\centering
   \includegraphics[width = 0.9\linewidth]{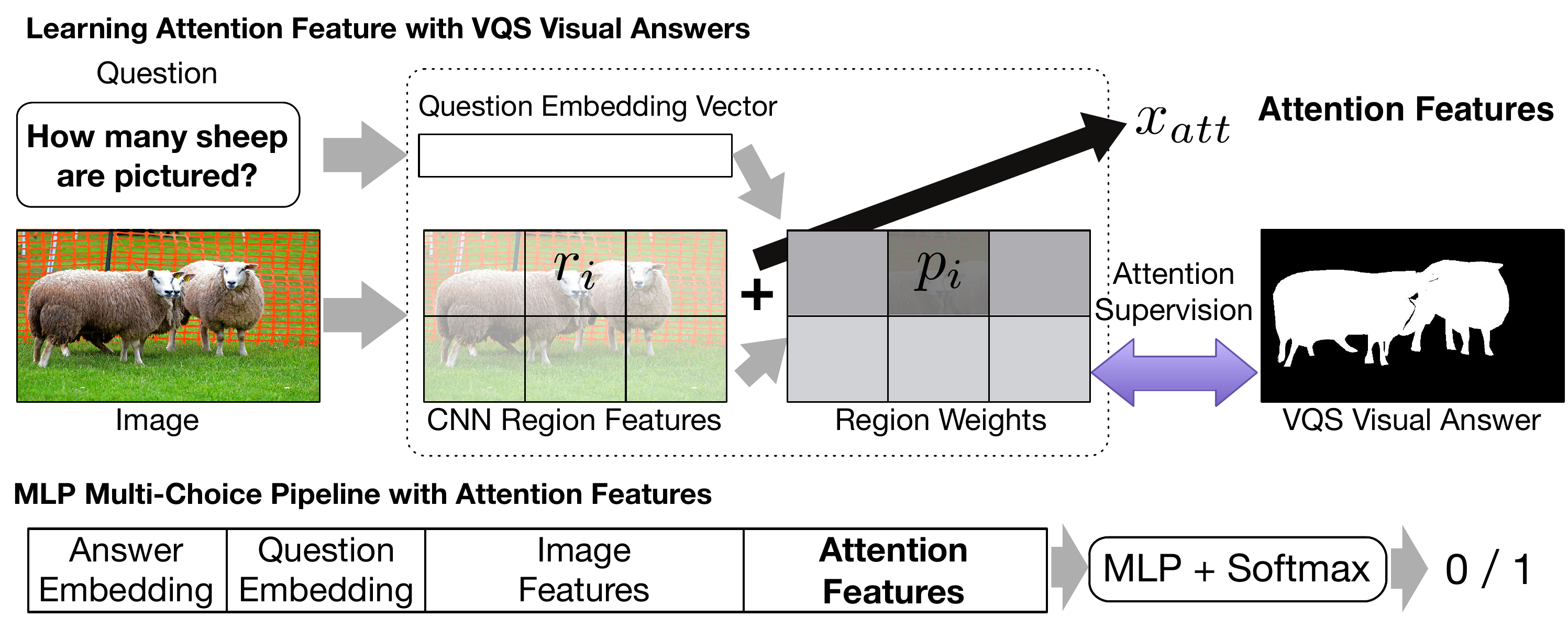}

   \caption{Supervised attention for VQA. To learn the attention features for each question-image pair, we use the corresponding segmentation mask as supervision to train the attention network. After that, we augment the MLP model~\cite{Revisiting_vqa} by the attention features.}
   \label{fig:supervised-vqa}
   \vspace{-10pt}
\end{figure*}

\subsection{Data statistics}
After collecting the annotations, we remove the question-image pairs for which the users selected the black buttons (full image) or gray buttons (unknown) to avoid trivial and ambiguous segmentation-QA links, respectively. In total, we keep 37,868 images, 96,508 questions, 108,537 instance segmentations, and 43,725 bounding boxes. In the following, we do not differentiate the segmentations from the bounding boxes for the ease of presentation and also for the sake that the bounding boxes are tight, small, and much fewer than the segmentations.

Figure~\ref{fig:segment} counts the distribution of the possible number of instance segmentations selected per image in response to a question. Over 70\% of questions are answered by one segmentation. On average, each question-image pair has 6.7 candidate segmentations, among which 1.6 are selected by the annotators as the visual answers.

In Figure~\ref{fig:distrbution}, we visualize the distribution of question types. The most popular type is the ``What'' questions (46\%). There are 31,135 ``is/are'' and ``does/do''  questions (32.1\%). Note that although the textual answers to them are simply yes or no, in VQS, we ask the participants to explicitly demonstrate their understanding about the visual content by producing the semantic segmentation masks. In the third column of Table~\ref{tab:mask}, we show the average number of segmentations chosen by the users out of the average number of candidates for each of the question types. 


\section{Applications of VQS} \label{sVQA}
The user linked visual questions and segmentations, where the latter visually answers the former, are quite versatile. They offer better leverage than before for at least two problems, i.e., supervised attention for VQA and question-focused semantic segmentation (QFSS).

\subsection{Supervised attention for VQA}
VQA is designed to answer natural language questions about an image in the form of short texts. We conjecture that a learning agent can produce more accurate text answers given the privileged access to the segmentations that are user linked to the QAs in training. To verify this point, we design a simple experiment to augment the MLP model in ~\cite{Revisiting_vqa}. The augmented MLP significantly improves upon the plain version and gives rise to state-of-the-art results on the VQA real multiple-choice task~\cite{VQA}.

\vspace{-4.5mm}
\paragraph{Experiment setup.}
We conduct experiments on the VQA Real Multiple Choices~\cite{VQA}. The dataset contains 248,349 questions for training, 121,512 for validation, and 244,302 for testing. Each question has 18 candidate answer choices and the learning agent is required to figure out the correct answer among them. We evaluate our results following the metric suggested in~\cite{VQA}.

\vspace{-4.5mm}
\paragraph{MLP for VQA Multiple Choice.}
Since the VQA multiple-choice task supplies candidate answers to each question, Jabri et al.\ propose to transform the problem to a stack of binary classification problems~\cite{Revisiting_vqa} and solve them by the multilayer perceptrons (MLP) model:
\begin{align}
y = \sigma(\bm{W}_2\,\max(0,\bm{W}_1\bm{x}_{iqa}) + b) \label{eMLP}
\end{align}
where $\bm{x}_{iqa}$ is the concatenation of the feature representations of an image, a question about the image, and a candidate answer, and $\sigma(\cdot)$ is the sigmoid function. The hidden layer has 8,096 units and a ReLU activation. This model is very competitive, albeit simple.

\subsubsection{Augmenting MLP by supervised attention} \label{sFeaturesVQA}
We propose to augment the MLP model by richer feature representations of the questions, answers, images, and especially by the supervised attention features detailed below.
\vspace{-8mm}
\paragraph{Question and answer features $\bm{x}_q \& \bm{x}_a$.} For a question or answer, we represent it by averaging the 300D word2vec~\cite{Mikolov2013Distributed} vectors of the constituent words, followed by the $l_2$ normalization. This is the same as in~\cite{Revisiting_vqa}.

\vspace{-4mm}
\paragraph{Image features $\bm{x}_i$.} We extract two types of features from an input image: ResNet~\cite{ResNet} pool5 activation and attribute features~\cite{wu2016value}, where the latter is the attribute detection scores. We implement an attribute detector by revising the output layer of ResNet. Particularly, given $C=256$ attributes, we impose a sigmoid function for each attribute and then train the network using the binary cross-entropy loss. The training data is obtained from the  COCO image captions~\cite{COCO}. We keep the most frequent 256 words as the attributes after removing the stop words.

\vspace{-5mm}
\paragraph{Attention features $\bm{x}_{att}$.} We further concatenate  attention features $\bm{x}_{att}$ to the original input $\bm{x}_{iqa}$. The attention features are motivated by the weighted combination of image regional features and question features in~\cite[eq.~(22)]{stacked}, where the the non-negative weight $p_i=f(Q,\{\bm{r}_i\})$ for each image region is a function of the question  $Q$ and regional features $\{\bm{r}_i\}$. We borrow the network architecture as well as code implementation from Yang et al.~\cite[Section 3.3]{stacked} for this function, except that we train this network by a cross-entropy loss to match the weights $\{p_i\}$ to the ``groundtruth'' attentions derived from the segmentations in our VQS dataset. In particular, we down-sample the segmentation map associated with each question-image pair to the same size as the number of image regions, and then $l_1$ normalize it to a valid probability distribution. By training the network to match the weights $p_i=f(Q,\{\bm{r}_i\})$ toward such attentions, we enforce larger weights for the regions that correspond to the user selected segmentations.


The upper panel of Figure~\ref{fig:supervised-vqa} illustrates the process of extracting the attention features, and the bottom panel shows the MLP model~\cite{Revisiting_vqa} augmented with our attention features for the VQA real multiple-choice task.

\begin{table}[t]
    \centering
 \caption{Comparison results on both VQA TestDev and Standard for the Real Multiple Choice task.}
\label{tab:vqa_choice}

    \begin{tabular}{c|c|c}

     Method &  Dev   & Standard  \\

     \hline

       Two-layer LSTM~\cite{VQA} & 62.7 & 63.1   \\

       Region selection~\cite{select_region} &62.4 & 62.4 \\

       DPPNet~\cite{dppnet} & 62.5 & 62.7 \\

       MCB~\cite{fukui2016multimodal}   & 65.4 & $-$ \\

       Co-Attention~\cite{coattention} & 65.9 & 66.1 \\

       MRN~\cite{MRN} &  66.3 &  66.3 \\

       MLB~\cite{kim2016hadamard}  &  $-$ & 68.9 \\

      \hline

       MLP + ResNet ~\cite{Revisiting_vqa}& 67.4 & $-$ \\
       MLP + ResNet +Atten. & 68.9 & $-$ \\

     \hline

       MLP + Attri. &  68.4 &  $-$ \\

       MLP + Attri. + Atten. & \textbf{69.5} & \textbf{69.8} \\

       \hline
       10 ensemble models  &  \textbf{70.5} &  \textbf{70.5} \\
       \hline

\end{tabular}
\vspace{-6mm}
\end{table}

%
%
%
%
%
%
%
%
%
%
%
%
%
%
%
%
%
%
%
%
%
%
%
%
%
%
%
%
%

\vspace{-3mm}
\subsubsection{Experimental results}
Table~\ref{tab:vqa_choice} reports the comparison results of the attention features augmented MLP with several state-of-the-art methods on the VQA real multiple-choice task. We mainly use the Test Dev for comparison. After determining our best single and ensemble models, we also submit them to the evaluation server to acquire the results on Test Standard.


First of all, we note that there is an 1.5\% absolute improvement over the plain MLP model (MLP + ResNet) by simply augmenting it using the learned attention features (MLP + ResNet + Atten.). Second, the attribute features for the images are actually quite effective. We gain 1.0\% improvement over the plain MLP by replacing the ResNet image features with the attribute features (cf.\ the row of MLP + Attri.\ vs.\ MLP + ResNet). Nonetheless, by appending attention features to MLP + Attri., we can still observe 1.1\% absolute gain. Finally, with an ensemble of five MLP + ResNet + Atten.\ models and five MLP + Attri.\ + Atten.\ models, our submission to the evaluation server was ranked to the second on Test Standard for the VQA real multiple-choice task, as of the paper submission date.


\vspace{-3mm}

\subsubsection{What is good supervision for attention in VQA?}
In this section, we contrast the VQS data to the human attention maps (HAT)~\cite{humman_attention} and bounding boxes that are placed tightly around the segmentations in VQS. The comparison results, reported in Table~\ref{tab:vqa_ablated}, are evaluated on the TestDev dataset of VQA Real Multiple Choice. We can see that the segmentaitons linked to QAs give rise to a little better results than bounding boxes, which further outperform HAT. These confirm our conjecture that HAT might be suboptimal for the supervised learning of attentions in VQA, since they reveal usually small parts of objects and contain large proportions of background. However, we believe it remains interesting to examine VQS  for more generic attention-based VQA models~\cite{stacked,xu2015ask,xiong2016dynamic,coattention,li2016visual,Andreas2016Neural,Multi_World}. 
\vspace{-4mm}
\paragraph{In the supplementary materials,} we describe the detailed implementation for the ensemble model. We also present additional results studying how different resolutions of the segmentation masks influence the VQA results.


\begin{table}[t]
    \centering
    \small
 \caption{Comparison results on VQA  TestDev Real Multiple Choice task. }
\label{tab:vqa_ablated}

    \begin{tabular}{c|c|c|c|c}

     Method &  Y/N & Num. & Others & All \\

     \hline

     Plain MLP~\cite{Revisiting_vqa}& 80.11 &38.88 & 64.17 &  67.49 \\
    \hline
     HAT~\cite{humman_attention} & 80.19 &39.34 & 64.92 &  68.42 \\


           Bounding boxes &  80.15 & 38.9 & 65.54 & 68.65\\
      \hline

    VQS  & \textbf{80.60} & \textbf{39.41} & \textbf{65.73} & \textbf{68.94} \\



       \hline
\end{tabular}
\vspace{-6mm}
\end{table}

\begin{figure*}
   \includegraphics[width = 1\linewidth]{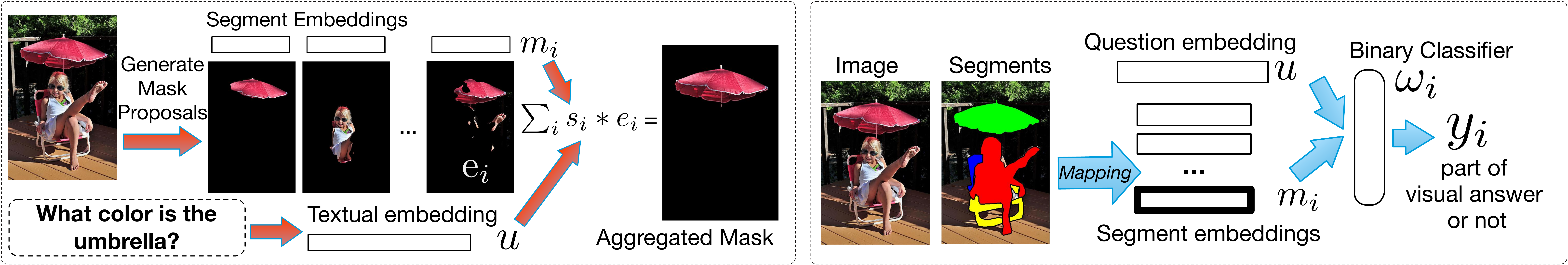}
   \centering
   \caption{Mask aggregation method for QFSS (left) and the method to estimate its upper bound performance (right).}
   \label{fig:maskagg}
   \vspace{-15pt}
\end{figure*}
\begin{figure}
   \includegraphics[width = 1\linewidth]{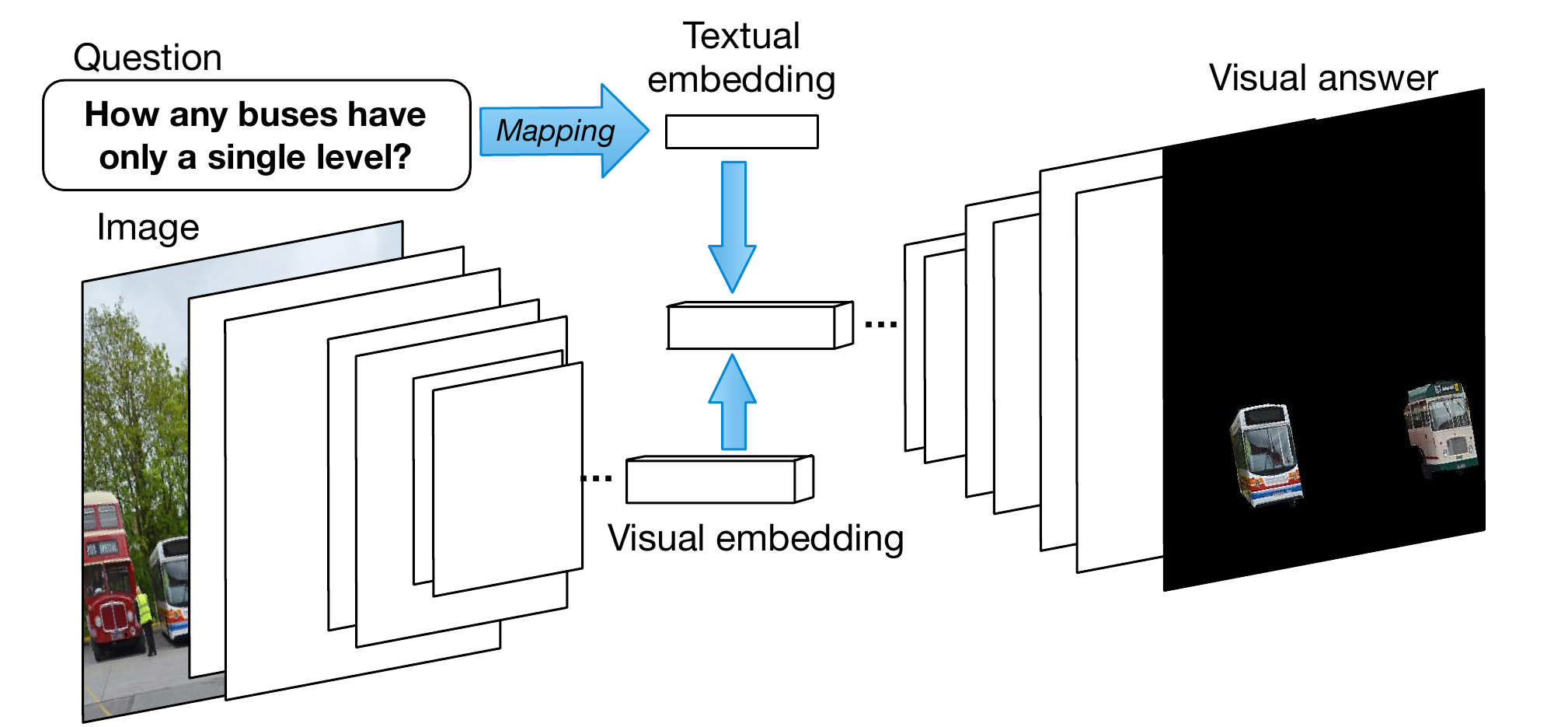}
\vspace{-5mm}
   \caption{Our DeconvNet baseline for QFSS.}
   \label{fig:DeconvNet}
\vspace{-4mm}
\end{figure}

\subsection{Question-focused semantic segmentation}
\label{sQFSS}
This section explores a new task, question-focused semantic segmentation (QFSS), which is feasible thanks to the collected VQS that connects two previously separate  tasks (i.e., segmentations and VQA). Given a question about an image, QFSS expects the learning agent to output a visual answer by semantically segment the right visual entities out of the image. It is designed in a way similarly to the segmentation from natural language expressions~\cite{segmentation_gronding}, with possible applications to robot vision, photo editing, etc.

In order to put the new task in perspective, we propose a mask aggregation approach to QFSS, study a baseline, and also investigate an upper bound  method by assuming all instance segmentations are given as oracles at the test stage.

\vspace{-3mm}
\subsubsection{Mask aggregation for QFSS}
We propose a mask aggregation approach to tackling QFSS. The modeling hypothesis is that the desired output segmentation mask can be composed from high-quality segmentation proposals. In particular, we use $N=25$ segmentation proposals $ e_{1}, e_{2}, \cdots, e_{N}$ generated by SharpMask~\cite{sharpmask} given an image. Each proposal is a binary segmentation mask of the same size as the image.

We then threshold a convex combination of these masks $E = \sum_i s_i e_i$ as the final output in response to a question-image pair, where the $i$-th combination coefficient $s_i$ is determined by the question features $\bm{x}_q$ and the representations $\bm{z}_i$ of the $i$-th segmentation proposal through a softmax function, i.e., $s_i=\text{softmax}(\bm{x}_q^TA\bm{z}_i)$. We learn the model parameters $A$ by minimizing an $l_2$ loss between the the user selected segmentations $E^\star$ and the model generated segmentation mask $E$. Our current model is ``shallow'' but it is straightforward to make it deep, e.g., by stacking its output with the original input following the prior practice (e.g., memory network~\cite{xiong2016dynamic} and stacked attention network~\cite{stacked}). 

\vspace{-4mm}
\paragraph{An oracle upper bound.}
We devise an upper bound  to the proposed method by 1) replacing the segmentation proposals with all the instance segmentations released by MS COCO, assuming they are available as oracles at testing, and 2) using a binary classifier to determine whether or not an instance segmentation should be included into the visual answer. The results can be considered an upper bound for our approach because the segmentations are certainly more accurate than the machine generated proposals, and the binary classification  is arguably easier to solve than aggregating multiple masks. We re-train the MLP (eq.~\ref{eMLP}) for the binary classifier here; it now takes as input the concatenated features of a segmentation and a question.

Figure~\ref{fig:maskagg} depicts the proposed approach and the upper-bound method with a concrete question-image example. 

\begin{table*}[t]
    \centering
\small
  \caption{Comparison results on QFSS (evaluated by IOU, the higher the better). For the question  representations, we consider the bag-of-words features (B) and the word embedding based features (W).}
\label{tab:mask}
\vspace{0.5em}
    \begin{tabular}{ccc|cc|cc|c}

          \hline
    Type &  Num. & \#seg ans/candts\ & Aggre. (B) & Aggre. (W)& DeconvNet (B) & DeconvNet (W)  & Upper  \\
    \hline
    All	   &14875	& 1.6/6.1  	& $\mathbf{0.3256}$ 	&	0.3174 	&	0.2687 	&	$0.2979$ &	   0.5709\\
    \hline
    does/do	& 561	& 1.6/6.0  & 	0.3294 	&	$\mathbf{0.3321}$ 	&	0.2751 	&	0.3297	&   0.5346\\
    how many	& 1814	& 2.2/6.3  & 	$\mathbf{0.3697}$ 	&	0.3645 	&	0.3147 	&	0.3370 	 &  0.6394\\
    is/are	&   4238	& 1.7/5.9  	&   $\mathbf{0.3672}$	&	0.3573 	&	0.3061 	&	0.3548 & 	0.6169\\
    what color & 1631 & 1.1/6.0  & 	$\mathbf{0.2596}$ 	&	0.2568 	&   0.1940 	&	0.1677 	 & 0.5557\\
    what is	&   2464 & 1.3/5.9 & 	$\mathbf{0.2472}$ 	&	0.2328 	&	0.2030 	&	0.2003 &  0.4987\\
    what (other) & 2722  & 1.6/6.1 & 	$\mathbf{0.3332}$ 	&	0.3235 	&	0.2556 	&	0.2809 	& 0.5482\\
    where	& 433	& 1.4/6.2  & 	0.1996 	&	$\mathbf{0.2040}$ 	&	0.1716 	&	0.1896 &   0.5707	\\
    which	& 202	& 1.4/5.9  & 	$\mathbf{0.2419}$ 	&	0.2339 	&	0.1695 	&	0.2012 & 0.4504	\\
    who	&	144	& 1.3/5.9   & 	$\mathbf{0.2573}$ 	&	0.2527 	&	0.2004 	&	0.2164 	& 0.2912 \\
    why	&  124 & 1.9/6.3  & 	0.3453 	& $\mathbf{0.3594}$ &	0.2430 	&	0.2917 	&   0.4781\\
    others&	542	& 1.6/6.1  & 	$\mathbf{0.3578}$ 	&	0.3354 	&	0.3097 	&	0.3534 &  0.5267	\\
\hline
\end{tabular}
\vspace{-12pt}
\end{table*}

\vspace{-4mm}
\paragraph{A baseline using deconvolutional network.}
Finally, we study a competitive baseline  which is motivated by the text-conditioned FCN~\cite{segmentation_gronding}. As Figure~\ref{fig:DeconvNet} shows, it contains three components, a convolutional neural network (CNN)~\cite{alexnet}, a deconvolutional neural network (DeconvNet)~\cite{Deconvolution}, and a question embedding to attend the feature maps in CNN. All the images are resized to $224\times 224$. The convolutional and deconvolutional nets follow the specifications in~\cite{Deconvolution}. Namely, a VGG-16~\cite{VGGNet} is trimmed till the last convolutional layer, followed by two fully connected layers, and then mirrored by DeconvNet. For the input question, we use an embedding matrix to map it to the same size as the feature map of the last convolutional layer. The question embedding is then element-wsie multiplied with the feature map. We train the network with an $l_2$ loss between the output mask and the groundtruth segmentation mask. 

\begin{figure}
   \centering
   \includegraphics[width = 0.9\linewidth]{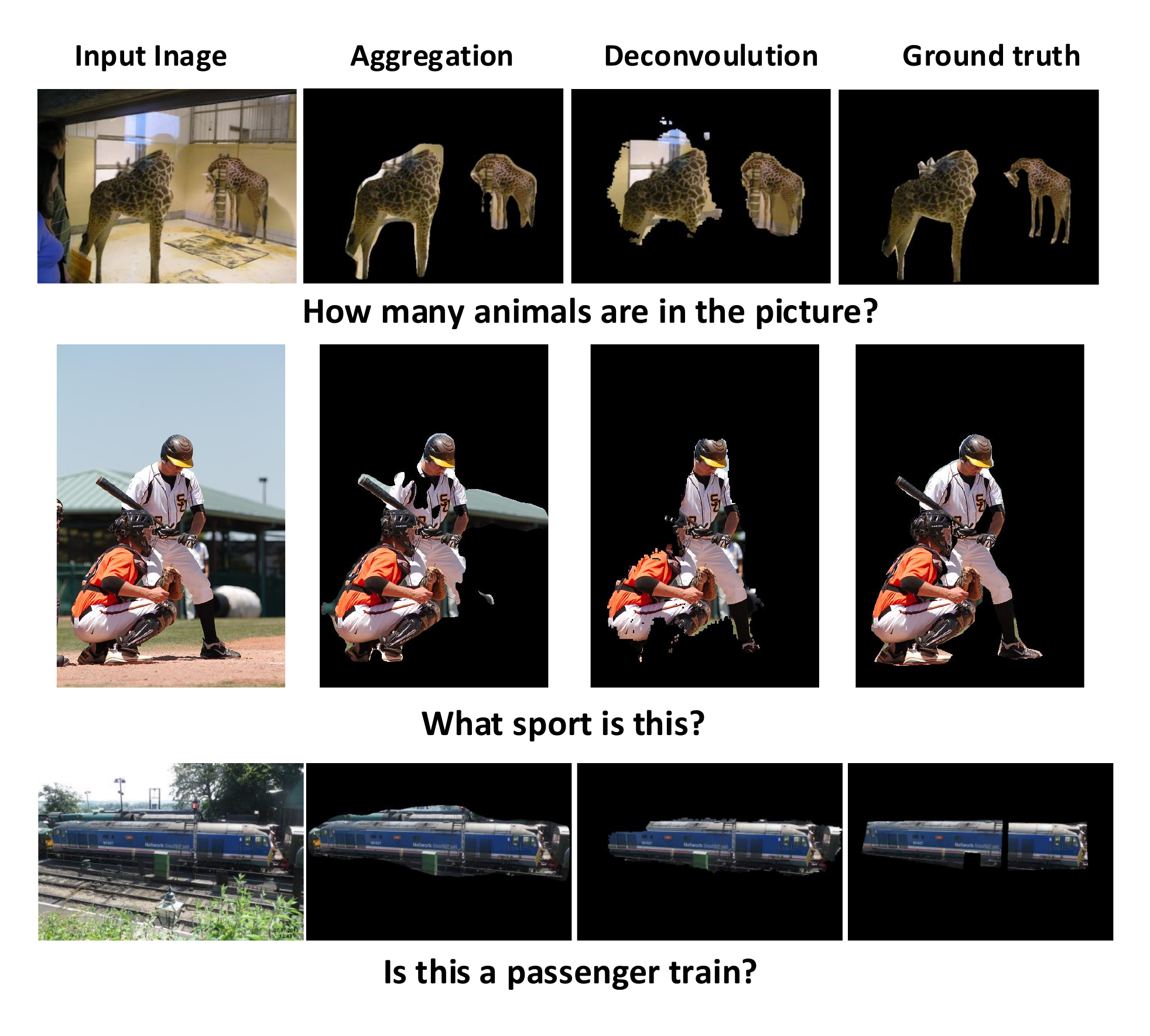}
   \vspace{-12pt}
   \caption{Qualitative results of mask aggregation and DeconvNet.}
   \label{fig:segmenatation}
   \vspace{-15pt}
\end{figure}

\subsubsection{Experiments on QFSS}

\vspace{-2mm}
\paragraph{Features.} In addition to representing the questions using the word embedding features $\bm{x}_q$ as in Section~\ref{sFeaturesVQA}, we also test the bag-of-words features. For each instance segmentation or proposal, we mask out all the other pixels in the image with 0's and then extract its features from the last pooling layer of a ResNet-152~\cite{ResNet}. 

\vspace{-5mm}
\paragraph{Dataset Split.}
The SharpMask we use is learned from the training set of MS COCO. Hence, we split our VQS data in such a way that our test set does not intersect with the training set for SharpMask. Particularly, we use 26,995 images and correspondingly 68,509 questions as our training set. We split the remaining images and questions to two parts: 5,000 images and associated questions for validation, and 5,873 images with 14,875 questions as the test set.

\vspace{-5mm}
\paragraph{Results.}
Table~\ref{tab:mask} reports the comparison results on QFSS, evaluated by intersection-over-union (IOU). In addition, the first three columns are about the number of different types of questions and the average numbers of user selected segmentations per question type. On average, more than one segmentations are selected for any of the question types.

First of all, we note that the proposed mask aggregation outperforms the baseline DeconvNet and yet is significantly worse than its upper bound method. The mask aggregation is superior over DeconvNet partially because it has actually used extra supervised information beyond our VQS data; namely, the SharpMask  is trained using all the instance segmentations  in the training set of MS COCO. The upper bound results indicate there is still large room for the mask aggregation framework to improve; one possibility is make it deep in the future work.

Besides, we find that the two question representations, bag-of-wrods (B) and word embedding (W), give rise to distinguishable results for either mask aggregation or DeconvNet. This observation is intriguing since it implies that the QFSS task is responsive to the question representation schemes. It is thus reasonable to expect that QFSS will both benefit from and advance the progress on joint vision and language modeling methods.

Finally, Figure~\ref{fig:segmenatation} shows some qualitative segmentation results. Note the two separate instance segmentations in the first row that visually answer the ``How many'' question.



\section{Conclusion}
\label{conclusion}
In this paper, we propose to link the instance segmentations provided by  COCO~\cite{COCO} to the questions and answers in VQA~\cite{VQA}. The collected links, named  visual questions and segmentation answers (\textbf{VQS}),
transfer human supervision between the individual tasks of semantic segmentation and VQA, thus enabling us to study at least two problems with better leverage than before: supervised attention for VQA and a novel question-focused semantic segmentation task. For the former, we obtain state-of-the-art results on the VQA real multiple-choice task by simply augmenting multilayer perceptrons with some attention features. For the latter, we propose a new approach based on mask aggregation. To put it in perspective, we study a baseline method and an upper-bound method by assuming the instance segmentations are given as oracles.

Our work is  inspired upon observing the popularity of  COCO~\cite{COCO}. We suspect that the existing and seemingly distinct annotations about MSCOCO images are inherently connected. They reveal different levels and perspectives of human understandings about the same visual scenes. Explicitly  linking them up  can significantly benefit not only individual tasks but also the overarching goal of unified vision-language understanding. This paper just scratches the surface. We will explore more types of annotations and richer models in the future work.
\vspace{-4mm}
\paragraph{Acknowledgement}
This work is supported in part by the NSF award IIS \#1566511, a gift from Adobe Systems Inc., and a GPU from NVIDIA. C.\ Gan is partially supported by the National
Basic Research Program of China  2011CBA00300 \& 2011CBA00301, and the National Natural Science Foundation of China  61033001 \& 61361136003.

\newpage
\section*{Appendices}

\appendix
\renewcommand{\appendixname}{Appendix~\Alph{section}}

\section{Annotation Interface}
\begin{figure*}[t]
   \centering
   \includegraphics[width=0.9\linewidth]{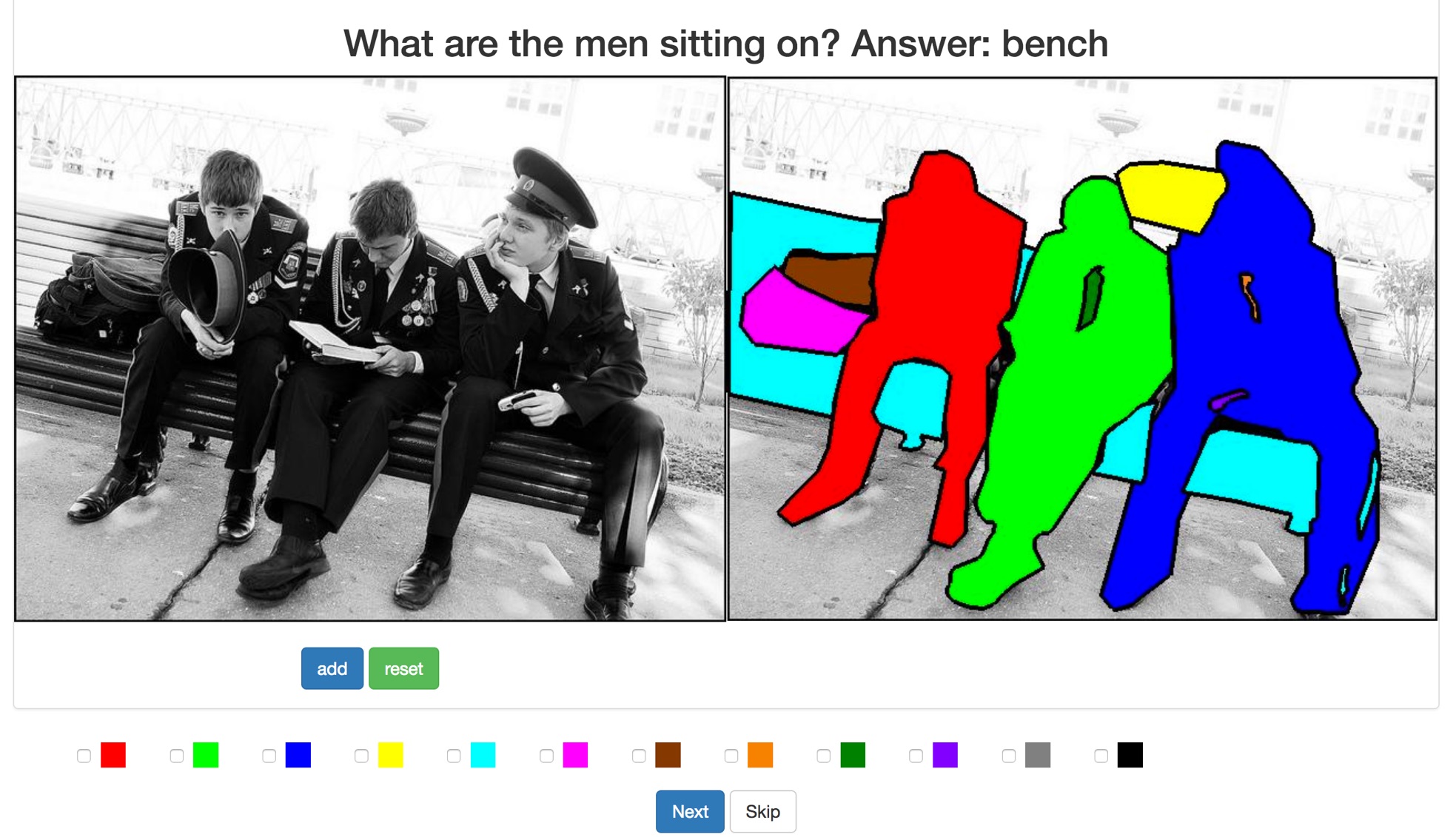}
   \caption{GUI we used to collect the links between image segmentations to questions and answers (VQS).}
   \label{fig:annotationUI}
\end{figure*}

Figure~\ref{fig:annotationUI} shows the annotation user interface we used to collect the VQS dataset.  Given a question about an image, the participants are asked to tick the colors of the corresponding segmentations to visually answer the question. The participants can also click the ``Add" button to draw bounding box(es) over the image in order to answer the question, in addition to choosing the segments. For more information please see the attached slides which we used to train the annotators.

\section{VQS \emph{vs.} VQA-HAT}
\begin{figure}
   \centering
   \includegraphics[width=0.9\linewidth]{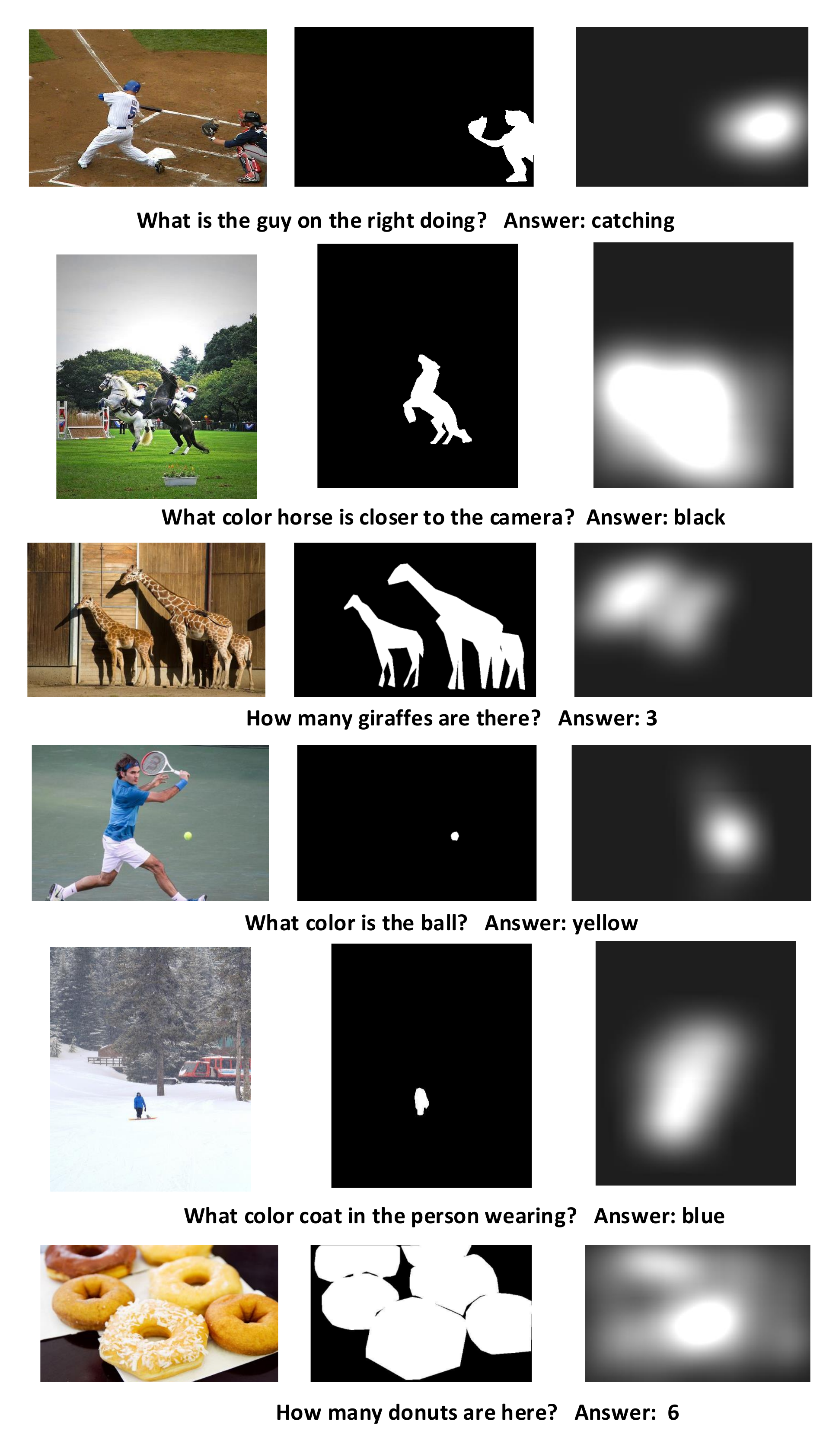}
   \caption{Comparing the segmentation annotations we collected for VQS with the human attention maps in VQA-HAT~\cite{humman_attention}.}
   \label{fig:compared}
\end{figure}

Figure~\ref{fig:compared} contrasts the human attention maps in VAQ-HAT~\cite{humman_attention} with our collected image segmentations that are linked by the participants to the questions and answers.  We observe that the HAT maps are rough comparing to the segmentation masks. For example, to answer the question ``what color is the ball?", our VQS dataset will provide a very accurate segmentation mask of the ball without including any background. We expect that such accurate annotations are more suitable for visual grounding tasks. Moreover, while segmentation is the desired final output in VQS,
the HAT maps mainly serve to analyze and potentially improve
VQA models that output/choose text answers.

\section{The influence of VQS segmentation mask resolution on the supervised attention in VQA}
The attention features we studied in Section 3.1.1 of the main text weigh the feature representations of different regions according to the question about the image. The number of regions per image indicate the attention resolutions. The more regions (the higher resolution) we consider, the more accurate the attention model could be. Of course, too small regions would also result in trivial solutions since the visual cues in each region would be too subtle then.

In the Table~\ref{tab:vqa_seg}, we report the VQA Real Multiple-Choice results on the Test-Dev by using different resolutions of the segmentation masks. We can observe that higher resolution leads to better VQA results. In some spirit, this implies the necessity of the accurate segmentation annotations for the supervised attention in VQA.

\begin{table}[ht]
    \centering
 \caption{Comparison results of segmentation mask resolutions for supervised attention in VQA.}
\label{tab:vqa_seg}

    \begin{tabular}{c|c|c|c|c}
    Method & Y/N & Num. & Others & All\\
   \hline
    VQS (14 $\times$ 14) & \textbf{80.60} & \textbf{39.41} & \textbf{65.73} & \textbf{68.94} \\
   \hline
    VQS (11 $\times$ 11) & 80.18 & 38.93 & 64.9 &  68.36 \\
    \hline
    VQS (7 $\times$ 7) & 79.49 & 38.08 & 63.71 &  67.41 \\

       \hline
\end{tabular}
\end{table}

\begin{table}
    \centering
 \caption{Comparison results of different language embeddings for VQS.}
\label{tab:vqS_new}

    \begin{tabular}{c|c|c}
   \hline
     DeconvNet (B) & DeconvNet (W) & DeconvNet (L)\\
    \hline
     0.2687 	&	0.2979 & \textbf{0.3144} \\
   \hline
\end{tabular}
\end{table}

\section{Some implementation details in the VQA and VQS experiments}

We use an ensemble of 10 models in our experiments for the VQA Real Multiple-Choice task (cf.\ Table 1 of the main text). Among them, five are trained using the attribute feature representations of the images and the other five are based on the ResNet features. We use the validation set to select the best 10 models as well as how to combine them by a convex combination of their decision values. After that, we test the ensemble on Test-Dev and Test-Standard, respectively.

For the VQS experiments, we use the ADAM~\cite{kingma2014adam} gradient descent to train the whole network with the learning rate 0.001 and batch size 16. It takes about one week on one Titan X GPU machine to converge after 15 epochs. We also report some additional results in Table \ref{tab:vqS_new} for our exploration of the LSTM language embedding in the DeconvNet approach. We observe that the LSTM language embedding model (L) gives rise to about 0.02 improvement over the bag-of-words (B) and word2vec embedding (W) on the challenging VQS task.

\end{document}